\documentclass[conference]{IEEEtran}
\IEEEoverridecommandlockouts
\usepackage{cite}
\usepackage{amsmath,amssymb,amsfonts}
\usepackage{algorithmic}
\usepackage{graphicx}
\usepackage{textcomp}
\usepackage{xcolor}
\usepackage{algorithm}
\usepackage{booktabs}

\usepackage{algorithm}
\usepackage{algorithmic}
\usepackage{booktabs}
\usepackage{amsmath} 
\usepackage{amsfonts}

\def\BibTeX{{\rm B\kern-.05em{\sc i\kern-.025em b}\kern-.08em
    T\kern-.1667em\lower.7ex\hbox{E}\kern-.125emX}}
\begin{document}

\title{Gated Graph Attention Networks for Predicting Duration of Large Scale Power Outages Induced by Natural Disasters}

\author{
\IEEEauthorblockN{
Chenghao Duan\IEEEauthorrefmark{1},
Chuanyi Ji\IEEEauthorrefmark{1},
Anwar Walid\IEEEauthorrefmark{2},
Scott Ganz\IEEEauthorrefmark{3}
}
\IEEEauthorblockA{\IEEEauthorrefmark{1}School of Electrical and Computer Engineering, Georgia Institute of Technology, Atlanta, GA, USA\\
Email: cduan8@gatech.edu, jichuanyi@gatech.edu}
\IEEEauthorblockA{\IEEEauthorrefmark{2}Department of Electrical Engineering, Columbia University, New York, NY, USA\\
Email: aie13@columbia.edu}
\IEEEauthorblockA{\IEEEauthorrefmark{3}School of Business, University of California, Riverside, Riverside, CA, USA\\
Email: scott.ganz@ucr.edu}
}

\maketitle

\begingroup
\renewcommand\thefootnote{}
\footnotetext{© 2026 IEEE. Personal use of this material is permitted. Permission from IEEE must be obtained for all other uses, in any current or future media, including reprinting/republishing this material for advertising or promotional purposes, creating new collective works, for resale or redistribution to servers or lists, or reuse of any copyrighted component of this work in other works. Accepted for publication in: Proceedings of the 2026 IEEE PES General Meeting, 18 – 21 January 2026, Montréal, Canada}
\addtocounter{footnote}{-1}
\endgroup

\begin{abstract}

The occurrence of large-scale power outages induced by natural disasters has been on the rise in a changing climate. Such power outages often last extended durations, causing substantial financial losses and socioeconomic impacts to customers. Accurate estimation of outage duration is thus critical for enhancing the resilience of energy infrastructure under severe weather. We formulate such a task as a machine learning (ML) problem with focus on unique real-world challenges: high-order spatial dependency in the data, a moderate number of large-scale outage events, heterogeneous types of such events, and different impacts in a region within each event. To address these challenges, we develop a Bimodal Gated Graph Attention Network (BiGGAT), a graph-based neural network model, that integrates a Graph Attention Network (GAT) with a Gated Recurrent Unit (GRU) to capture the complex spatial characteristics. We evaluate the approach in a setting of inductive learning, using large-scale power outage data from six major hurricanes in the Southeastern United States. Experimental results demonstrate that BiGGAT achieves a superior performance compared to benchmark models. 

\end{abstract}


\section{Introduction} \label{Intro}

Natural disasters such as hurricanes, winter storms, wildfires, and flooding have become more frequent and intense in a changing climate \cite{IPCC, Emanuel_natcomm_2022}. Energy infrastructure is particularly vulnerable, as severe weather events have induced widespread power outages in the US and worldwide \cite{stankovic2022methods, adhikari2025quantifying}. Enhancing the resilience for the energy infrastructure has been advocated by both research communities and industry, where a pertinent aspect of resilience is to reduce interruption duration from power outages \cite{xu2024resilience}. In this context, accurately predicting outage duration is important for emergency preparation and response, resource allocation, and infrastructure planning \cite{ganz2023socioeconomic, afsharinejad2021large}.

We view outage duration prediction as a machine learning (ML) problem: given large-scale power outages induced by prior natural disasters, learn a mapping between outage durations and feature variables on weather, outages, geo-spatial and demographic characteristics. Importantly, we aim to evaluate how accurately the mapping can predict power outage durations from an impending weather event. This task presents three real-world challenges:  (a) Spatial dependency: Severe weather induced outage durations and feature variables at one location are correlated with those in neighboring and possibly distant areas \cite{best2023spatial}. (b) Spatial heterogeneity: Weather severity and outages vary spatially, with durations ranging from days near the storm track to hours at locations farther away \cite{afsharinejad2021large, ji2016large}. (c) Moderate data: Historical data on severe weather-induced outage events is either rare or not publicly available. The data scarcity makes it challenging to train a model that generalizes to new events in different locations.

The objective of this work is to study these challenges to (1) Develop a graph neural network as the mapping, through a novel combination of Graph Attention Network (GAT) \cite{velivckovic2017graph}  with Gated Recurrent Unit (GRU) \cite{li2015gated, cho2014learning}, which captures spatial heterogeneity and dependency in large-scale power outages. (2) Evaluate the model’s ability to generalize to unseen events in a setting of inductive learning using field data.

\begin{figure}[!h]
    \centering
    \includegraphics[width=0.8\linewidth]{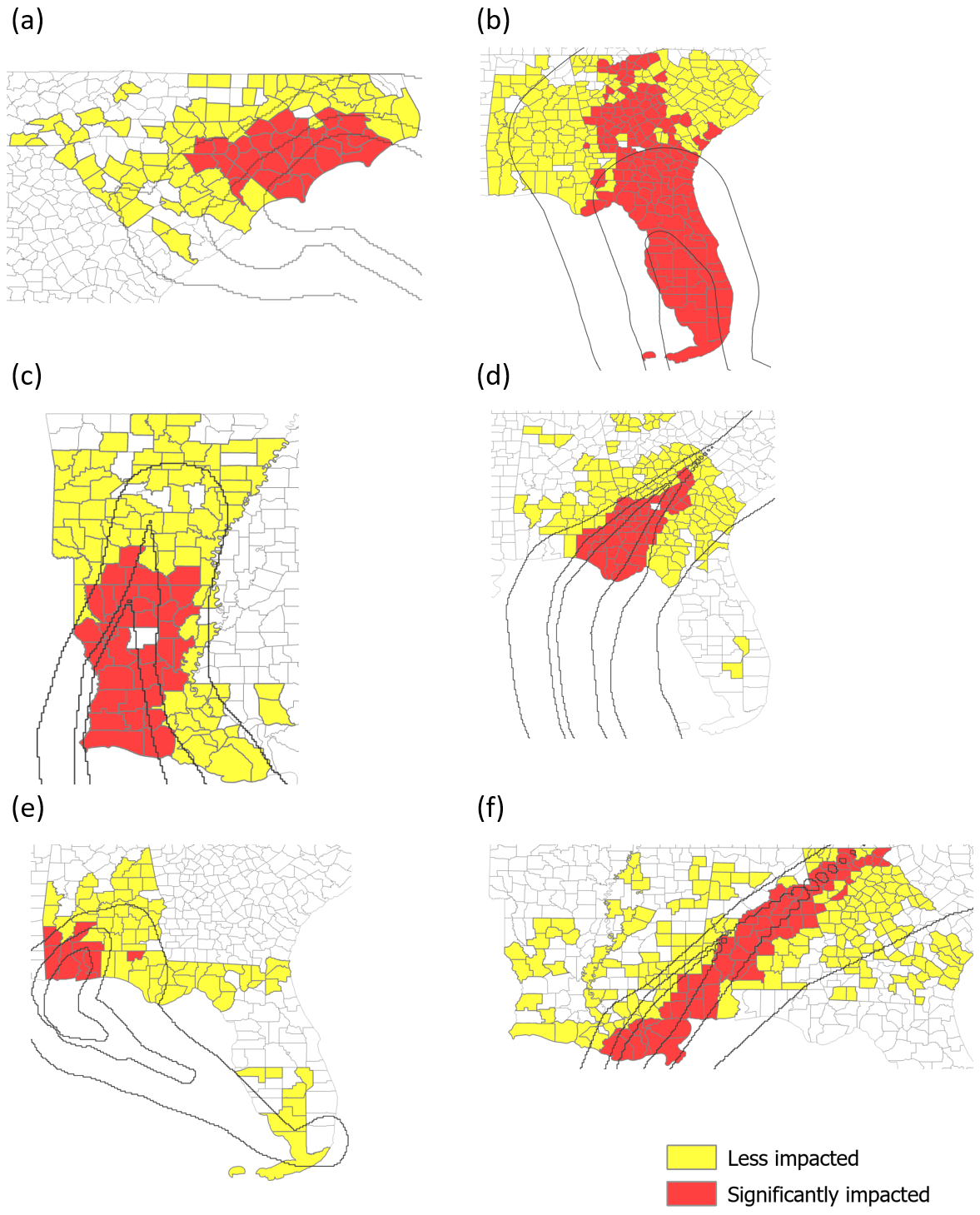}
    \caption{Spatial heterogeneity and dependency of power outages from six hurricanes: (a) Florence, (b) Irma, (c) Laura, (d) Michael, (e) Sally, (f) Zeta. Dark lines: wind boundaries (64, 50, and 34 knots). Clusters on regions with different impacts:  Red - significantly impacted counties; Yellow - less impacted counties. Clustered by wind and outage duration.} 
    \vskip -0.2in
    \label{fig:LauraCls}
\end{figure}
\subsection{Related work}

{\bf Statistical machine learning approaches on power outages:} Prior work on predicting severe weather-induced outage durations has focused on statistical and machine learning methods. Tree-based models achieve state-of-the-art performance for outage duration estimation \cite{nateghi2014forecasting, mcroberts2018improving, fang2022failure}. Neural networks \cite{yu2018deepsolar} are also used for the problem. These approaches, however, do not model spatial characteristics explicitly, and thus require large datasets on historical power outages \cite{fang2022failure}. When the data are limited, the studies are often restricted to small spatial scopes \cite{nateghi2014forecasting}. Spatial statistics have been applied to outage duration prediction \cite{ganz2023socioeconomic, best2023spatial}. These methods, however, generally assume homogeneous spatial correlation across locations \cite{lesage2009introduction}, and thus are unable to characterize spatial heterogeneity in weather-induced power outages.

{\bf Graph neural networks and the power grid:} Graph based models have been applied to power grid analysis but little to weather-induced large-scale outages using real data. In particular, most prior works focus on power flow prediction, load forecasting, and outage/attack detection \cite{liao2021review, dey2019network}. The prior works often rely on synthetic datasets. Field data have mainly been used for outage detection rather than duration prediction \cite{owerko2018predicting}. 

Overall, the research challenges described in the introduction are open and unique for learning from real-world natural disaster-induced large-scale power outage using field data.

\section{Method}
We now describe the development of BiGGAT model. We first present variables and datasets, then problem formulation, algorithm design, and experiment setting. 

\subsection{Variables}
We select the outage duration as the output variable $\boldsymbol{Y}$ of BiGGAT model. Here the outage duration $\boldsymbol{Y}$ characterizes the impact of severe weather-induced large-scale power failures. The feature variables $\boldsymbol{X}$ include (a) hurricane wind swath as an exogenous cause of power outages, (b) the maximum number of affected customers at given locations, (c) geo-spatial and demographic characteristics of affected regions including spatial adjacency, population density and area, and (d) socioeconomic vulnerability of affected communities. 

\subsection{Field data}\label{DurThres}
The power outage data we obtain are from the top six major Atlantic hurricanes that made landfall between 2017 and 2021 in the Southeastern United States (Fig. \ref{fig:LauraCls}) \cite{ganz2023socioeconomic}. These events caused outages across 557 counties and affected 15 million customers. County-level, time-varying outage data from PowerOutage.com \cite{powerUS22} provides the number of customers without service. Outage duration of an affected county is defined as the time from the maximum  number of disrupted customers to when this number falls below 5\% of customers served.

Weather variables are sourced from the National Hurricane Center’s best-track data \cite{HURDAT2}, specifying hurricane wind swaths at each affected location. Maximum sustained wind speeds are sorted into three categories (34-49 knots, 50-63 knots, and 64+ knots). Population, area, and spatial adjacency are obtained from the US Census Bureau \cite{USCensusBureau2020}. Socioeconomic vulnerability is characterized by the CDC/ATSDR Social Vulnerability Index (SVI) \cite{flanagan2018measuring}, which is widely used in prior works \cite{ganz2023socioeconomic}.

In practice, outage impact is evaluated by duration levels rather than precise length, e.g., whether outages are short (lasted for hours) or prolonged (for days) \cite{GPEAP24}. Thus, we label outage durations to three classes, following Department of Energy (DOE)'s restoration guidelines \cite{doe2022playbook}: short ($<$ 2 days), medium (2–6 days), and long ($>$ 6 days). Outage labels correspond to our model output $\boldsymbol{Y}$. The labeled dataset is summarized in Table \ref{Datasummary}.
\vskip -0.1in
\begin{table}[!h]
\caption{Data summary: number of counties for three classes of outage durations}
\vskip -0.2in
\label{Datasummary}
\begin{center}
\begin{small}
\begin{sc}
\begin{tabular}{lcccr}
\toprule
Events & Short & Medium & Long \\
\midrule
Florence  & $68$ & $21$ & $9$ \\
Irma      & $158$ & $127$ & $29$ \\
Laura     & $76$ & $18$ & $17$ \\
Michael   & $127$ & $31$ & $18$ \\
Sally     & $62$ & $7$ & $1$ \\
Zeta      & $179$ & $44$ & $12$\\
\bottomrule
\end{tabular}
\end{sc}
\end{small}
\end{center}
\vskip -0.1in
\end{table}

\subsection{Problem formulation}
We formulate the problem as node-level classification using the graph neural network, that is, learning a mapping from input features $\boldsymbol{X}$ to outage-duration labels $\boldsymbol{Y}$ via a graph structure. We make the following assumptions for our formulation: 
\begin{enumerate}
\item Spatial dependency is considered up to $n^{th}$ order geo-spatial neighbors with $n = 1$ for the nearest and $n>1$ for farther neighbors. 
\item Spatial dependency is modeled within each event. Cross-event dependencies are not considered in  this work.
\item County-level variables are used due to data availability.
\end{enumerate}

We let $G = \{V, E, \boldsymbol{A}\}$ represent the graph structure of the outage data, where $V$ is a set of affected counties (nodes). $E$ consists of neighboring county pairs (edges).  $\boldsymbol{A}$ is a symmetric adjacency matrix, with $\boldsymbol{A}_{ij}=1$ if nodes $i$ and $j$ are neighbors, $0$ otherwise. For each node $v \in V$ in the data, $\boldsymbol{x}_{v} \in  \mathbb{R}^{11}$ represents an input feature vector, including the maximum number of affected customers, county population, county area, four variables for social vulnerability and four for hurricane wind swaths. $\boldsymbol{X}=\{\boldsymbol{x}_v\}$ for all $v\in V$ represents the set of all node features. $\boldsymbol{y}_{v} \in \mathbb{R}^{3}$ denotes the node-level 3-class duration labels (short, medium and long), as defined in Field Data. The problem is formalized as a graph learning task, i.e., to obtain a GNN model $M$ that predicts class labels of outage duration $\boldsymbol{Y}$ using node features $\boldsymbol{X}$ and graph structure $G$, 
\begin{align*}
    \boldsymbol{Y} &= M(\boldsymbol{X}, G).
\end{align*}

\subsection{Bimodal gated graph attention network}
We develop a GNN with two key components: (a) a Bimodal Embedding and (b) a Gated Graph Attention mechanism. The Bimodal Embedding is established to capture heterogeneity through unsupervised learning. The Gated Graph Attention mechanism provides a novel integration of self-attention \cite{velivckovic2017graph} with GRU \cite{li2015gated, cho2014learning} for message propagation.



\subsubsection{Bimodal graph node embedding}
For each node $v \in V$, Bimodal Embedding maps the input $\boldsymbol{x}_v$ to a message, $\boldsymbol{m}_{v,k}$ for $k=0$, where $k$ is the number of recurrent iterations for the Gated Graph Attention layer. Through unsupervised learning (K-means algorithm), the data on wind swath and duration labels form two spatial clusters as illustrated by Fig. \ref{fig:LauraCls}. The distinct clusters characterize the heterogeneous impact of a weather event. The embedding structure is determined accordingly by the number of clusters. For instance, the two clusters for this case result in two distinct sets of learnable weights, $\boldsymbol{\beta}^{(l_v)}$ ($l_v\in\{1,2\}$). This results in a simple linear embedding, where for a given node $v \in V$, the embedded message $\boldsymbol{m}_{v,0}$ is
\begin{align*}
\boldsymbol{m}_{v,0} &= \boldsymbol{\beta}^{(l_v)} \boldsymbol{x}_{v},
\end{align*}
where $\boldsymbol{m}_{v,0}$ is a three-dimensional vector representing the initial message at iteration $k = 0$. The number of clusters can vary to represent the complexity of the model and spatial heterogeneity. 

\subsubsection{Gated graph attention mechanism}

\begin{figure*}[!htbp]
\centering
\includegraphics[width=0.75\linewidth]{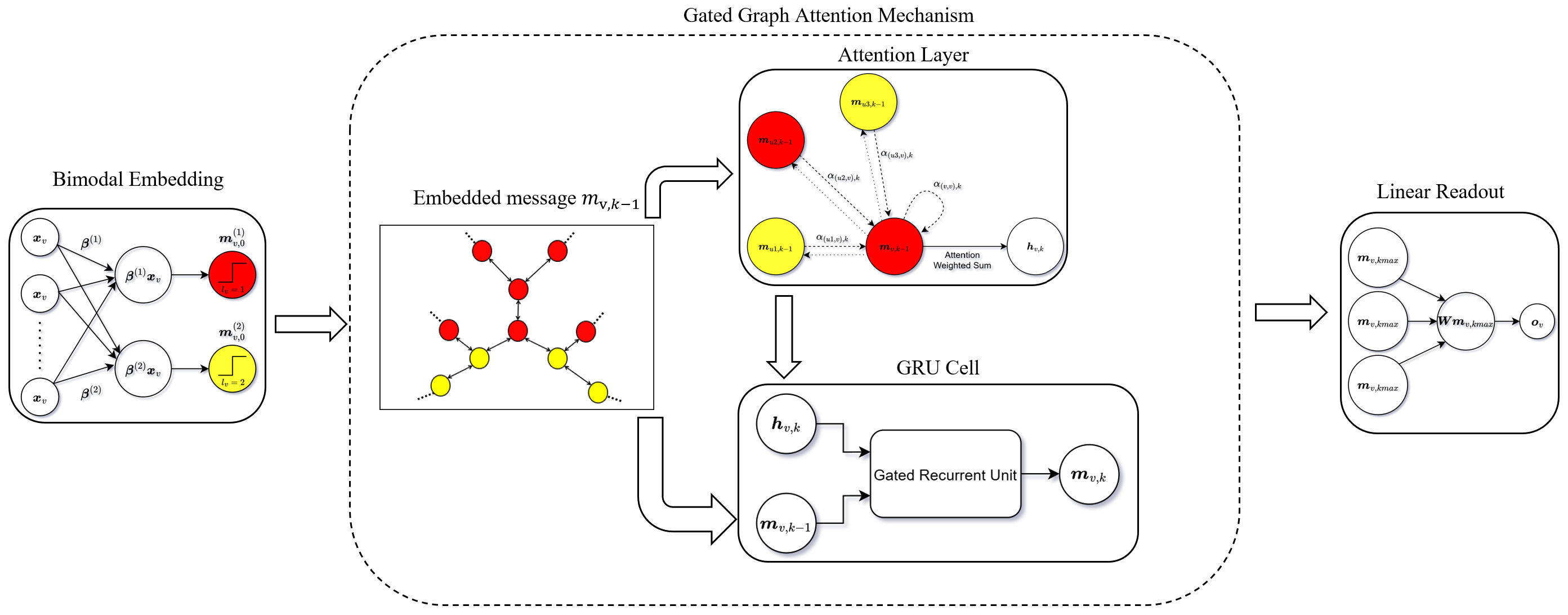}
\caption{Overall BiGGAT structure: Node features are fed into the Bimodal Embedding, then passed through the Gated Graph Attention mechanism. The aggregated messages are transformed to outputs by a Linear Readout layer.} 
\vskip -0.2in

\label{fig:BiGRANNet}
\end{figure*}

Gated Graph Attention mechanism combines a GRU with self-attention. In particular, after the Bimodal Embedding layer, each node’s initial message $\boldsymbol{m}_{v,0}$ is aggregated via the attention layer into an intermediate embedding $\boldsymbol{h}_{v,1}$ of the same dimension as $\boldsymbol{m}_{v,0}$. The attention layer characterizes the spatial dependency weights. A GRU cell then propagates messages to higher-order neighbors with memory: at each iteration $k \in [1,k_{\text{max}}]$, the GRU updates $\boldsymbol{m}_{v,k}$ from the intermediate embedding $\boldsymbol{h}_{v,k}$ and the previous message $\boldsymbol{m}_{v,k-1}$. Here $kmax$ is a hyperparameter of design. The basic message propagation step is:
\begin{align*}
    \boldsymbol{h}_{v,k} &= Attn(\boldsymbol{m}_{v,k-1}, N_v),\\
    \boldsymbol{m}_{v,k} &= GRU(\boldsymbol{m}_{v,k-1},\boldsymbol{h}_{v,k}),
\end{align*}
where $Attn$ refers to the self-attention based graph message aggregation \cite{velivckovic2017graph} among the neighbors $N_v$, and $GRU$ refers to the Gated Recurrent Unit \cite{cho2014learning, li2015gated}. 

The embedded message $\boldsymbol{m}_{v,k}$ is fed into the Gated Graph Attention mechanism for multiple iterations to propagate the influence of each node to its higher-order neighbors. After a given number $k_{max}$ of iterations, the output message will be passed to a linear readout layer to translate into output labels $\boldsymbol{o}_{v}$ at each node. Fig. \ref{fig:BiGRANNet} illustrates the overall structure of the BiGGAT model. The overall pipeline of the BiGGAT can be summarized as in Algorithm \ref{alg:BiGRAN}
\vskip -0.1in
\begin{algorithm}[!h]
\caption{Overall BiGGAT pipeline}\label{alg:BiGRAN}
1. Curate graph-structured outage data using geo-spatial statistics.\\
2. Use K-means to determine the structure of the Bimodal Embedding layer. \\
3. Embed node message $\boldsymbol{m}_{v,0}$ using the Bimodal Embedding layer.\\
4. Use the self-attention mechanism to update the hidden embedding $\boldsymbol{h}_{v,k}$.\\
5. Feed both the previous embedded message $\boldsymbol{m}_{v,k-1}$ and the current hidden embedding $\boldsymbol{h}_{v,k}$ into GRU for the current embedded message $\boldsymbol{m}_{v,k}$.\\
6. Repeat step 4 and 5 for $k \in [1,kmax]$.\\
7. Feed the embedded message from step 6 to a linear read out layer for prediction $\boldsymbol{o}_{v}$. Evaluate the loss and accuracy.\\
\vskip -0.1in
\end{algorithm}

\subsection{Experiment setting}

We adopt the setting of inductive learning to apply BiGGAT to real-world scenarios. We further specify the experiment designs and describe three metrics for performance evaluation. 

\subsubsection{Inductive learning}

The setting of inductive learning is desirable for real-world implementation where a model is trained on historical events and then predicts the outage durations of an unseen event. To implement such a setting, we hold out one event as the testing set and use the other five events to train BiGGAT. For test samples, the cluster labels $l_v$ are inferred using input features. 

One key design for the model is the graph structure of the data - characterized by the neighborhood $N_v$ of each node. To capture sufficient spatial dependency, we define $N_v$ to include all counties up to the highest spatial order of significant correlation. Meanwhile, for simplicity, we define $N_v$ to include neighbors up to the same spatial order $n$ for all nodes within a given event. The actual value of $n$ is obtained from data: we measure the spatial correlation of the feature variable, peak outages, using the global Moran's I \cite{moran1950notes}. The Moran's I, a widely-used spatial statistic, measures the weighted covariance of the variables among the immediate geo-spatial neighbors. We extend the measure to farther spatial neighbors by adapting the spatial weight matrix $A$ in the Moran's I calculation with an n-hop neighbor adjacency matrix $\boldsymbol{A}_{spatial}^{(n)}$:
\begin{align*}
    \boldsymbol{A}_{spatial}^{(n)} = I\{\boldsymbol{A}_{spatial}^{n} > 0\} - I\{\boldsymbol{A}_{spatial}^{n-1} > 0\}
\end{align*}

Where $\boldsymbol{A}_{spatial}$ denotes the spatial adjacency matrix of affected counties and $I$ denotes an indicator function. Thus, $\boldsymbol{A}_{spatial}^{(n)}$ is nonzero only at entries where two given locations are exactly $n^{th}$ order neighbor to each other without redundancy. Such an n-hop Moran's I effectively captures the correlation among the n-hop spatial neighbors. 

\begin{table}[h]
\caption{Global n-hop Moran's I of peak outage number}
\vskip -0.2in
\label{tbl-morani}
\begin{center}
\begin{small}
\begin{sc}
\begin{tabular}{lcc}
\toprule
Events & Highest n & Global n-hop Moran's I \\
\midrule
Florence  & $3$ & $0.273^{*} (6.806)$  \\
Irma      & $6$ & $0.070^{*} (4.723)$  \\
Laura     & $5$ & $0.112^{*} (3.499)$\\
Michael   & $3$ & $0.053^{*} (1.998)$\\
Sally     & $2$ & $0.110^{*} (1.864)$ \\
Zeta      & $4$ & $0.154^{*} (6.049)$\\
\midrule
\multicolumn{3}{p{0.4\textwidth}}{\small \textit{ $^*$ denotes rejection of null hypothesis with $p\leq0.1$. Values in brackets are z-scores. }}
\end{tabular}
\end{sc}
\end{small}
\end{center}
\vskip -0.3in
\end{table}

The n-hop Moran’s I (Table \ref{tbl-morani}) shows significant positive spatial correlation for at least the second-order geo-spatial neighbors. Therefore, we select the highest significant n-hop number to define the size of neighborhood $N_v$. In addition, we choose the max recurrence iteration number $kmax$ to be $2$. Thus, with control from GRU, the messages are not only aggregated within neighborhood $N_v$ but also propagated to farther neighbors beyond $N_v$. This design ensures that our model can fully capture the high-order spatial correlations present in each event.

Our model is trained for node-level multi-class classification. Three metrics are used to benchmark the performances: (1) Classification Accuracy, (2) Macro F1 Score, and (3) Balanced Accuracy. We use the cross-entropy loss and ADAM optimizer for training, which are implemented using Pytorch, PyGeometrics, and Sklearn packages via Google Colab.

\section{Results}

\subsection{Inductive learning model performance}
The BiGGAT and benchmark models, including XGBoost (XGB), Rnadom Forest (RF), GAT, and BiGAT, are trained on five events and tested on the remaining unseen event. These experiments are repeated six times, with each event serving as the test set once. For each model, the corresponding hyperparameters are iteratively tuned to the corresponding optimal performances.

\begin{figure}[!h]
    \centering
    \includegraphics[width=0.9\linewidth]{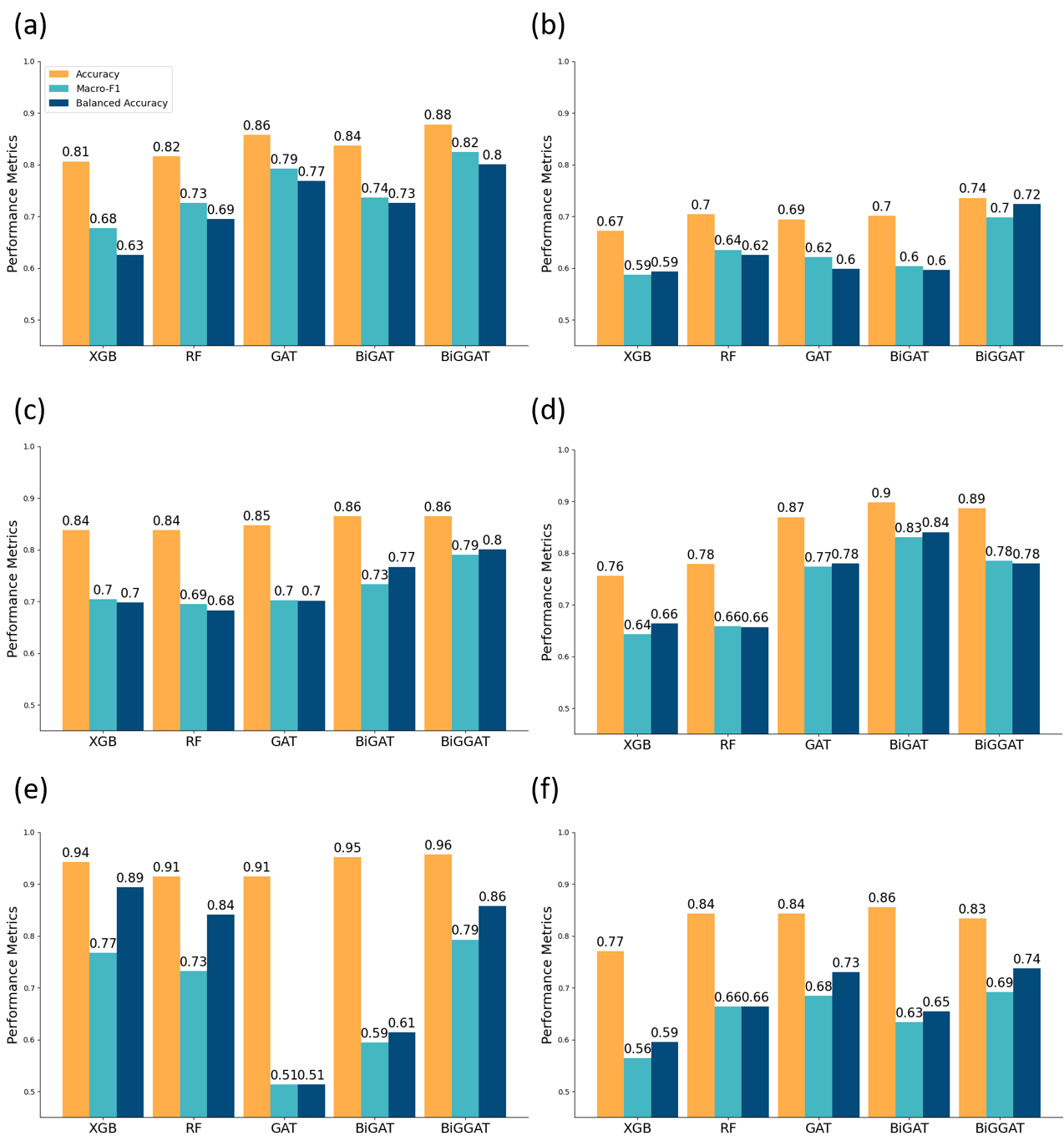}
    \caption{Model performance comparison for (a) Florence, (b) Irma, (c) Laura, (d) Michael, (e) Sally, and (f) Zeta. Each subplot shows classification accuracy, Macro F1, and balanced accuracy for XGB, RF, GAT, BiGAT, and BiGGAT.}
    \vskip -0.1in
    \label{fig:PerformanceResultsAll}
\end{figure}

\begin{table}[h]
\caption{Event Average performance of models}
\vskip -0.2in
\label{tbl-AvgPerf}
\begin{center}
\begin{small}
\begin{sc}
\begin{tabular}{lccc}
\toprule
Models & Accuracy  & Balanced Acc & Macro F1 \\
\midrule
XGB  & $79.7\%$ & $67.8\%$ & $0.657$ \\
RF      & $81.5\%$ & $69.4\%$ & $0.682$\\
GAT    & $83.7\%$ & $68.1\%$ & $0.678$\\
BiGAT   & $85.1\%$ & $70\%$ & $0.686$\\
BiGGAT    & $85.9\%$ & $78.3\%$ & $0.762$ \\
\midrule
\end{tabular}
\end{sc}
\end{small}
\end{center}
\vskip -0.2in
\end{table}

Event-wise comparison and averaged metrics are shown in Fig. \ref{fig:PerformanceResultsAll} and Table \ref{tbl-AvgPerf} respectively. BiGGAT consistently outperforms the tree-based models (XGB and RF), indicating that graph-based models better capture spatial dependencies in the outage data. Among graph-based models, BiGGAT and BiGAT outperforms GAT in five events, underscoring the value of modeling spatial heterogeneity via Bimodal Embedding. Compared with BiGAT, which uses only nearest neighbors and without GRU, BiGGAT achieves better class-balanced performance in all events. These improvements indicate better accuracies for medium- and long-duration classes in the imbalanced dataset. There are fewer medium and long duration samples in the data, which are more challenging to learn. This also highlights the importance of modeling high-order spatial dependencies. Overall, the BiGGAT shows significant improvements over existing models on learning outage durations in the challenging setting of inductive learning.

\subsection{Generalization of GRU and error analysis}

The inductive setting for outage duration prediction requires generalization across hurricanes affecting different regions. To evaluate this, we split test counties from unseen events into two groups: those with geographical overlap with the training set (``test with overlap'') and those without any geographical or structural connection (``absolute disjointed''). The latter group contains 265 test counties and represents the most challenging cases for generalization. Nonetheless, their test accuracy (82.4\%) is comparable to the overall test accuracy, and errors from these counties account for only 27.3\% of all test errors. These results indicate that BiGGAT generalizes well to unseen and structurally unconnected samples.

\section{Conclusion}

This work presents a novel application of graph neural networks for predicting duration of power outages induced by natural disasters. We develop BiGGAT that captures spatial heterogeneity and high-order spatial dependencies in the field data. BiGGAT learns from and is tested by six historical Atlantic hurricanes under an inductive learning setting. To achieve robustness with moderate event data, the model adopts a relatively simple architecture with  Bimodal Embedding, self-attention and GRU modules for message passing. Experimental results show that BiGGAT significantly outperforms benchmark models. 

Further, the Gated Graph Attention mechanism enhances the generalization performance and class-balanced metrics in inductive learning. The model also performs robustly on the test samples that are located in differently regions from the training data. These results underscore the importance of modeling higher-order spatial dependencies, particularly under severe weather events, and validate the use of GRU and self-attention for message passing.

There are several research directions for future study: (a) extending BiGGAT to other events and disaster types, including  wildfires, winter storms and tornadoes, (b) adapting the model to data with finer spatial resolution, e.g., private data owned by power utilities, (c) exploring how predicted outage impacts benefit vulnerability assessments and infrastructure. All these will benefit informed decision-making in disaster preparation and resilience strategies.

\bibliographystyle{IEEEtran}
\bibliography{BiGRAN}

\end{document}